\documentclass[12pt]{article}

\usepackage{amsmath}
\usepackage{amssymb}

\usepackage{graphicx}

\usepackage{cite}

\usepackage[labelfont=bf,labelsep=period,justification=raggedright]{caption}
\usepackage{caption}

\makeatletter
\renewcommand{\@biblabel}[1]{\quad#1.}
\makeatother

\date{}

\begin{document}

\begin{flushleft}
{\Large
\textbf{Niche as a determinant of word fate in online groups}
}
\\ \bigskip
Eduardo G.\ Altmann$^{1,2,3}$, 
Janet B.\ Pierrehumbert$^{1,4}$, 
Adilson E.\ Motter$^{1,5,\ast}$
\\ \medskip
{1} {\it Northwestern Institute on Complex Systems, Northwestern University, Evanston, Illinois, United States of America.}
\\
{2} {\it Departamento de F\'{i}sica, Universidade Federal do Rio Grande do Sul, Porto Alegre, Rio Grande do Sul, Brazil.}\\ \smallskip
{3} {\it Max Planck Institute for the Physics of Complex Systems, Dresden, Germany.}\\     \smallskip
{4} {\it Department of Linguistics, Northwestern University, Evanston, Illinois, United States of America.}\\ \smallskip
{5} {\it Department of Physics and Astronomy, Northwestern University, Evanston, Illinois, United States of America.}\\
\medskip
$\ast$ E-mail: motter@northwestern.edu.
\end{flushleft}

\section*{Abstract}

Patterns of word use both reflect and influence a myriad of human activities and interactions. Like
other entities that are reproduced and evolve, words rise or decline depending upon a complex interplay
between {their intrinsic properties and the environments in which they function}.
Using Internet discussion communities as model systems,  we 
define the concept of a {\em word niche} as the relationship between the word and the
characteristic features of the environments in which it is used. 
We develop a method to quantify two important aspects of the size of  the word niche: the
range of individuals using the word and the range of  topics it is used to discuss.
Controlling for word  frequency, we show that these aspects of the word niche are strong determinants of
changes in word frequency. Previous studies have already
indicated that word frequency itself is a correlate of word success at historical time scales. Our 
analysis of changes in word frequencies over time
reveals that the relative sizes of word niches are far more important than word frequencies in the
dynamics of the entire vocabulary at shorter time scales, as the language adapts to new concepts and
social groupings. We also distinguish endogenous versus exogenous factors as additional contributors to
the fates of words, and demonstrate the force of this distinction in the rise of novel words.
Our results indicate that short-term nonstationarity in word statistics is
strongly driven by individual proclivities, including inclinations to provide novel 
information and to project a distinctive social identity.\\

\noindent
{\bf Cite as:  PLoS ONE 6(5), e19009 (2011). doi:10.1371/journal.pone.0019009.}

\section*{Introduction}

Much information about the fabric of modern human society has been gleaned from large-scale
records of human communications activities, such as time stamps and network structures for email
exchanges, mobile phone calls, and Internet activity~\cite{onnela,Barabasi,malmgren,seshadri}. 
But the flow of words  has
the potential to be even more informative. Words characterize both external events  and otherwise
unobservable mental states. They tap into the variety of experience, knowledge, and goals
of different interacting individuals. The word stream is information-dense, because the number of distinct 
words and expressions is so great.  The lexicon of a literate adult is estimated to contain
over 100,000 distinct items~\cite{kuiper}, and it continues to grow as new words are
encountered~\cite{davis}.

Records of the linguistic transactions within a community provide an ongoing statistical sampling
of the vocabulary of a language. The sample at any time reflects both the social context
(who is speaking, and to whom) and the topical context (what they are speaking about). But the
language dynamics does not just passively mirror the context. Language adapts
to new circumstances and needs through lexical innovation~\cite{munat}. 
Large datasets available from the Internet provide an unprecedented opportunity to study the
dynamics of words, as well as phrases and tags~\cite{kleinberg, cattuto,hotho, neuman}. 
Here, we
explore lexical fluctuations in relation to both individuals and topics by analyzing records of Usenet
groups. Created over one decade before the World Wide Web, the Usenet groups were amongst the first
systems for world-wide exchange of messages on the Internet. Usenet archives reveal the rise of
``Netspeak", the language nowadays widely used on the Internet and in telephone text
messages~\cite{Crystal}. The  groups we studied, rec.music.hip-hop and comp.os.linux.misc, were
selected for their great lexical creativity. In these datasets, users serve as proxies for
individuals, and threads as proxies for topics (see {\em Methods}). Our study goes
beyond the analysis of user activity in Usenet groups~\cite{B1}, and focuses instead on the content of
the messages. 

It is known that word frequency is a factor in frequency dynamics on historical time 
scales~\cite{Pagel,Lieberman}, a finding that is expected from models of language learning across human
generations~\cite{fontanari}. Here, we identify two new factors---the dissemination of words across individuals (users)
and the dissemination of words across  topics (threads)---and we develop a method to quantify dissemination that  controls for word
frequency. Because words are acquired and reproduced by users as they communicate with each other about different
topics, these two dissemination measures serve to characterize two important dimensions of the word niche. We apply
these measures  to demonstrate that dissemination is  a much more powerful determinant of word fate than word
frequency is; poorly disseminated words are more likely to experience a frequency reduction than widely disseminated  words.

{These results suggest analogies between word fates and the fates of biological species. In population
biology, the term {\em niche} refers to the relationship between a species and the aspects of its environment
that enable it to live and reproduce. Quantifying the  breadth and versatility of a species' niche, as
distinct from the species' sheer abundance, is key to understanding its competitive position within an 
ecosystem~\cite{Colwell}. The geographic size of the niche is a statistical 
correlate of species duration, as species with large ranges are less likely to 
become extinct~\cite{foote, jablonski}. 
Analogies between language and population biology have proved fruitful in
understanding the dynamics of entire languages, in particular the relationship of community size to
overall rates of linguistic change~\cite{pop1,pop2} and to properties of the syntactic and morphological systems~\cite{pop3}. 
Here, we work at a more fine-grained level, quantifying the impact at short (two-year) time scales of
the heterogeneous usage of language inside a community. Because we consider the role of
heterogeneity amongst people within the community,
the results also support comparisons between the dynamics of the linguistic
system and other social dynamics, such as the spread of opinions or the
popularity of news items, videos, and music~\cite{watts, salganik}.}

The relation with social dynamics
is strengthened by a case study of novel words with rising frequency, in which we compare a set of words for products and public figures to a set of slang words.
The rise in use of words in the first set is mainly driven exogenously by events that are external to the Usenet group, such as product releases, 
political crises, and public performances.
Because the use of slang words is strongly influenced by the social
values and patterns of communication within any given linguistic group
~\cite{eble, smitherman}, the 
use of the (slang) words in the second set should be more influenced by factors endogenous to the Usenet community.
The force of this
distinction in word dynamics mirrors its force in other social behaviors,
ranging from YouTube viewing to scientific discoveries, marketing successes, financial
crashes, and civil wars~\cite{Sornette, Crane}. Finally,
we  explore the correlations between individuals and topics as dimensions of word dissemination.
The two dimensions are shown to be separable, and individual choices prove to be more important
than topic in determining patterns of word usage. These results highlight the importance of
individuality in the use of language, and imply limits on the role of 
social influence and social conformity.

\section*{Results}

\subsubsection*{Dissemination of words across users and threads}\label{ssec.dissemination}\ If everyone knew the
same words, and chose to use them at random with their given frequencies, the dissemination of words
across users would be the result of a Poisson process. We are interested in the
extent to which
the actual number of users of each specific word
deviates from this baseline model.
We define the measure of dissemination of each word~$w$ across users 
as
\begin{equation}\label{eq.dissemination}
D^U_w=\frac{U_w}{\tilde{U}(N_w)}, 
\end{equation}
where $N_w$ is the number of occurrences of the word in the dataset, 
$U_w$ is the actual number of users whose posts include word~$w$ at least once,
and~$\tilde{U}$ is the expected number of users predicted by the baseline model.
The latter is determined from $\tilde{U}=\sum^{N_U}_{i=1}\tilde{U}_i$, where
$N_U$ is the number of users and 
$\tilde{U}_i$  is the probability that user $i$ used $w$ at least once
when all the words in the text are shuffled randomly
(see {\em Methods}). 
Dissemination  across threads is analogously defined as
\begin{equation}\label{eq.disseminationT}
D^T_w=\frac{T_w}{\tilde{T}(N_w)}, 
\end{equation}
where $T_w$ is the number of threads in which the word appears, and $\tilde{T}$ is the 
corresponding expected value from the baseline model. { The word frequency is defined
as~$f=N_w/N_A$, where $N_A=\sum_w N_w$ is the total number of words in the dataset;} 
{$N_w$ is a count, and the frequency $f$ normalizes this count to a probability}.
In the rest of the paper, 
we focus on the properties of the {dissemination} measures~$D^U_w$ and $D^T_w$, or~$D^U$ and $D^T$ 
for notational simplicity.

\begin{figure*}[!ht] 
\begin{center} 
\includegraphics[width=0.9\columnwidth]{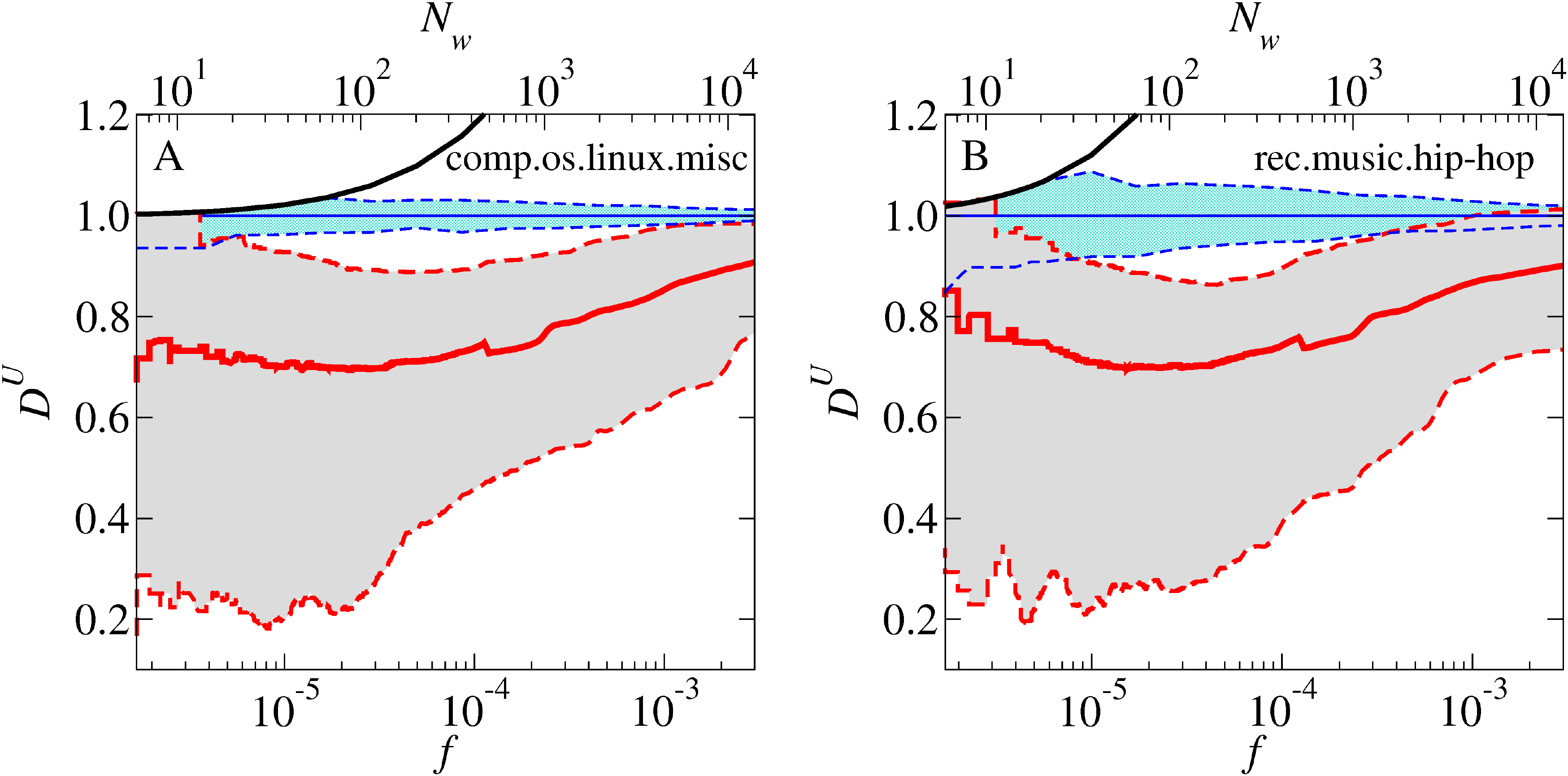}
\caption {\baselineskip 14pt
{\bf Relationship of frequency $f$ to dissemination {across} users $D^U$.}
{\bf A},{\bf B}, The results are shown for half-year windows centered on 1998-01-01 for
the comp.os.linux.misc group ({\bf A}) and the rec.music.hip-hop group ({\bf B}).
Red solid line: running median for all words with
$N_w > 5$. Red dashed lines: $10$th and $90$th percentiles for the same words.
Blue dashed lines:
$10$th and $90$th percentiles around the expected value of $D^U$ for randomly distributed words, 
determined by Monte Carlo simulations with $100$ independent shufflings of the text.
Black line: analytically calculated ceiling 
$D^U_{\max}=N_w/\tilde{U}$ (floor
effects and the other ceiling, $D^U_{\max}=N_U/\tilde{U}$, do not pertain within the scale of the figure).
The median empirical $D^U$ is systematically below the $10$th percentile of the estimated random variation.
The relationship of median $D^U$ to $f$ is nearly flat up to $\log_{10}f = -3.5$.}
\label{fig1}\label{figure1} 
\end{center}
\end{figure*}

The expected value of $D^U$ is $1$ for a word of any frequency that is distributed randomly across
users. $D^U>1$ indicates {\em over-disseminated} words and~$D^U<1$ indicates concentrated or {\em
clumped} words. For example, in a half-year window centered on 1998-01-01 in the
comp.os.linux.misc group, the words {\it thanks} and {\it redhat} have almost
identical frequencies, but contrast in their dissemination ({\it thanks}: $N_w = 4,121$, $D^U =
1.19$; {\it redhat}: $N_w = 4,146$, $D^U = 0.75$). A similar
contrast is provided for the same time window in the rec.music.hip-hop group by the words {\it please} ($N_w = 2,336$,
$D^U = 1.17$) and {\it article} ($N_w =  2,366$, $D^U = 0.59$). 
The measure $D^U$ exhibits a lower bound
determined by the number of occurrences of the word:~$\frac{1}{N_w} \le D^U$. For
any given set of posts, there is also an upper bound determined by the relationship of $N_w$ and
$N_U$ to $\tilde{U}$: $D^U \le \min\{N_w,N_U\}/\tilde{U}$. Due to the discreteness close to the
lower bound, we set a threshold $N_w > 5$ for the computation of $D^{U,T}$. The few dozen most
frequent words (mainly common function words) are also omitted from our analysis, because $D^U$ is
not informative when $N_w$ is too large compared to the number of users. Figure~\ref{fig1} shows
results on the expected statistical fluctuation around $D^U=1$ for randomly distributed words in a
representative window of each Usenet group, as determined by a Monte Carlo simulation.  The upper
and lower extremes of the fluctuation depend on frequency, but only slightly.

The dissemination across threads $D^T$ is closely related to  the {\em residual inverse
document frequency} ($r\mbox{-}IDF$), a measure used in text processing to characterize the extent to
which a word is associated with particular documents~\cite{church95,manning}. 
$IDF$, defined as the
reciprocal of the number of documents in which the word occurs, is strongly influenced by word
frequency. {\em Residual} $IDF$ addresses this artifact by taking the difference $r\mbox{-}IDF =
\log(\tilde{T}) - \log(T)$, where $\tilde{T}$ is approximated using a Poisson baseline model with
equal document lengths. When this condition holds,
$-\log(D^T) = r\mbox{-}IDF$. 
The measure $D^T$ is a generalization
of $r\mbox{-}IDF$ that remains valid when the lengths of the documents are very unequal, as for 
the present datasets (see Supporting Information S1, Figure~S1).

\subsubsection*{$D^U$ and $D^T$ as predictors of word fate}
\begin{figure*}[t!] 
\includegraphics[width=0.96\columnwidth]{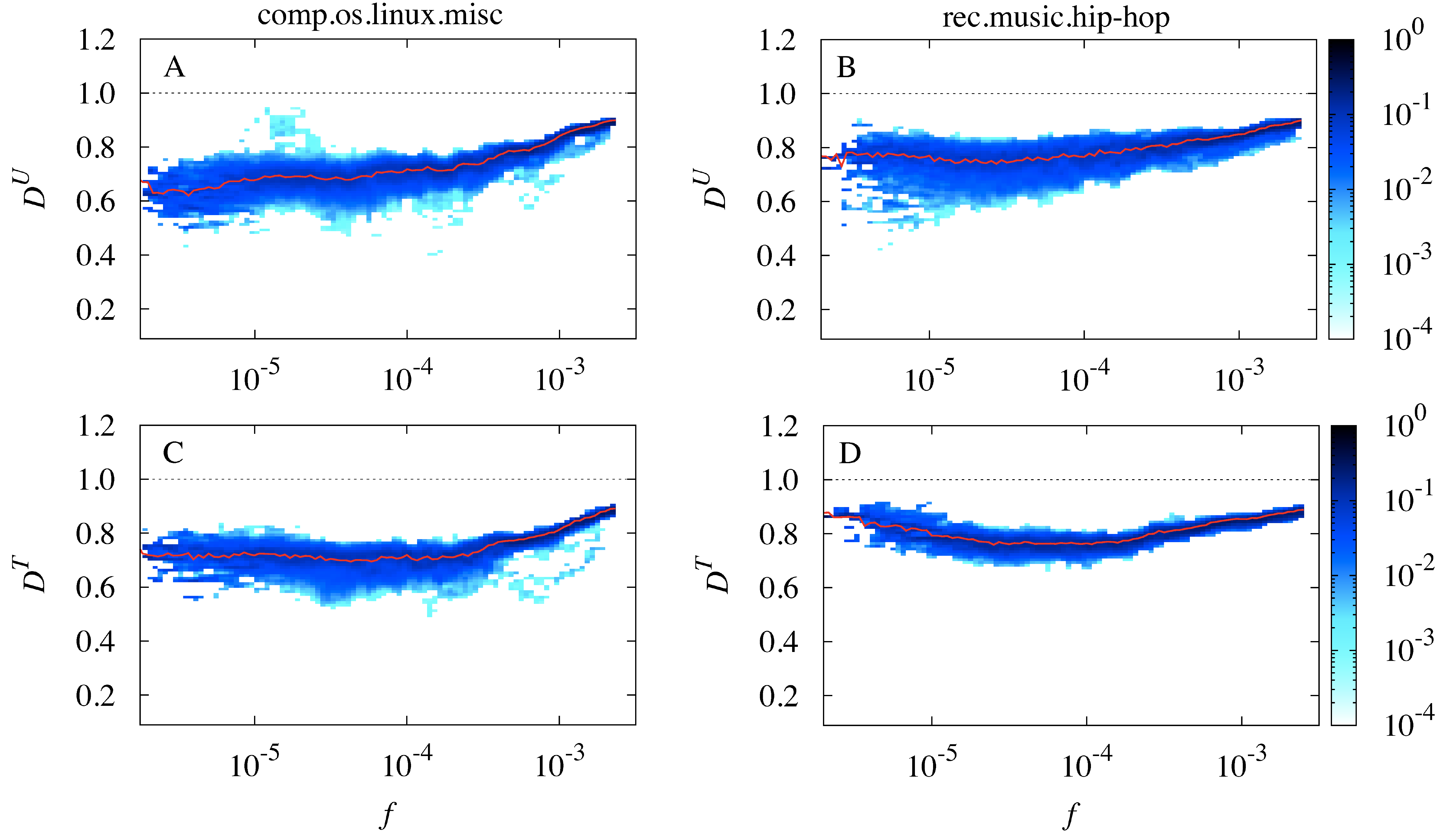}
\caption{ {\bf Summary of the relation between frequency $f$ and dissemination {across} users $D^U$ and threads~$D^T$.} 
The running median shown in Figure~\ref{fig1} is now calculated in all half-year windows.
{\bf A}-{\bf D}, Results for both the comp.os.linux.misc 
group ({\bf A},{\bf C})
and the rec.music.hip-hop group ({\bf B},{\bf D}).
The color code indicates densities in the range of $10^{-4}$ (light blue) to $1$ (dark blue)
obtained by combining all running medians, while the red line indicates the median of the resulting,
combined distribution.
}
\label{figure2}
\end{figure*}

\ 
To explore the changes over time in the statistical attributes of words, we begin by partitioning
each dataset into non-overlapping half-year windows. Figure~\ref{fig1} displays the behavior of $D^U$
within a representative half-year window for both groups. Most words are significantly clumped. At
all word frequencies, the median $D^U$ falls below the 10th percentile for random fluctuation of
the expected value under the baseline model. For words with $\log_{10}f < -3.5$,  $D^U$ varies
considerably and is not correlated with frequency $f$. Words with $\log_{10}f > -3.5$  are extremely
high-frequency words, and  comprise less than  0.5\% of all distinct words in this window. 
But even these words are somewhat clumped. These
findings are reproduced in all half-year windows for both Usenet groups, as summarized in
Figure~\ref{figure2}AB.
They provide the user counterpart to prior observations of clustering of words in
documents and in time~\cite{church95,manning,serrano, kleinberg, Altmann}. 

\begin{figure*}[!ht]  
\begin{center} 
\includegraphics[width=1.0\columnwidth]{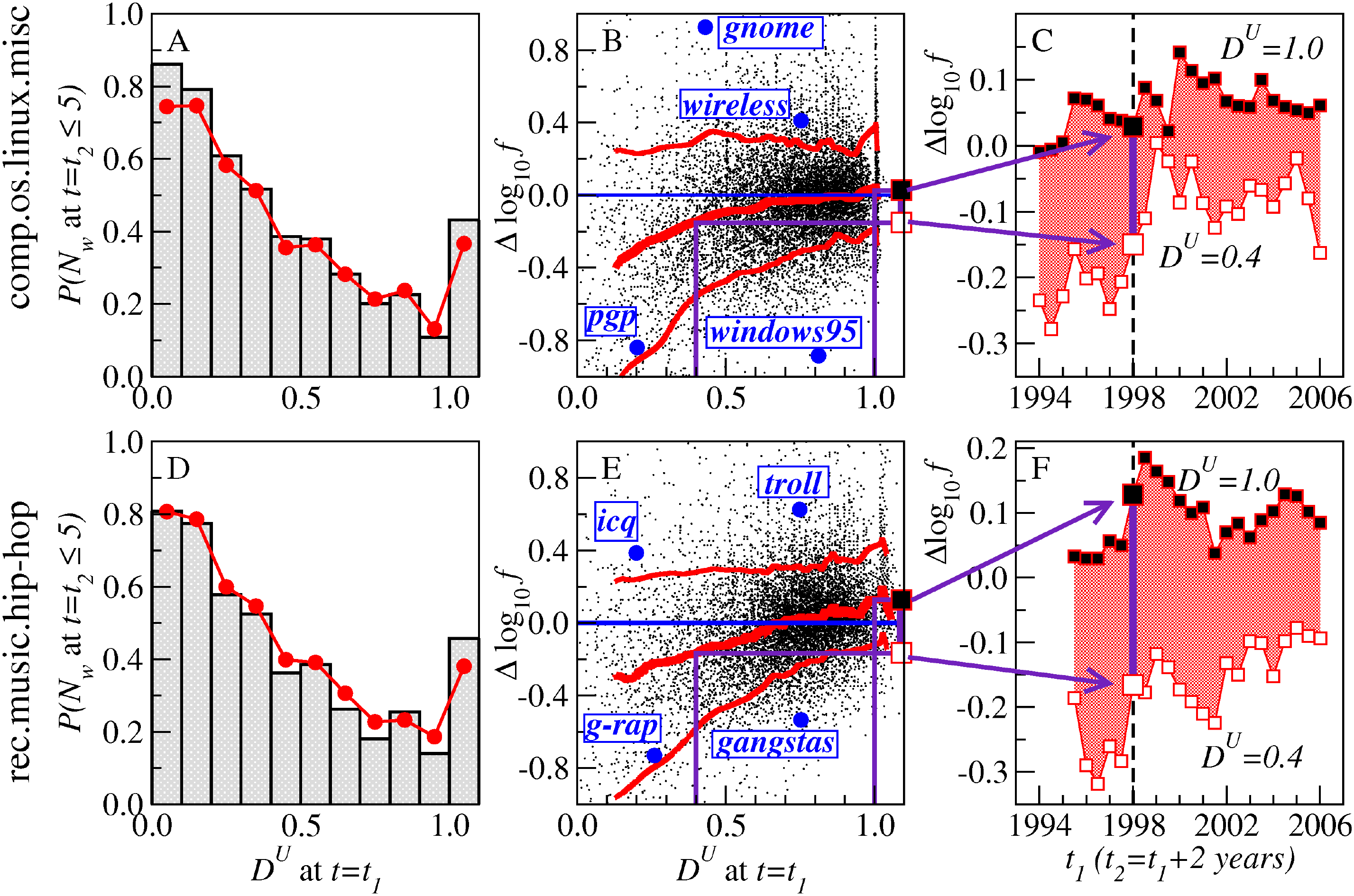}
\caption{\baselineskip 14pt
{\bf Dissemination {across} users $D^U$ as a predictor of falling below threshold and of frequency decay.}
The analysis is performed over half-year window pairs  $t_1$ and~$t_2$
separated by two years for the comp.os.linux.misc and rec.music.hip-hop groups.
{\bf A},{\bf D}, Fraction of words with~$N_w>5$  in~$t_1$ that fall to $N_w\leq5$ in $t_2$. 
Histogram in gray: results from selected window pairs centered 
on $t_1 = 1998\mbox{-}01\mbox{-}01$ and $t_2 = 2000\mbox{-}01\mbox{-}01$. 
Red line: average over different non-overlapping window pairs with $t_1$ ranging
from the (rounded off) beginning of the group 
through 2006-01-01, 
and $t_2=t_1+ 2$ years.
The probability of falling below threshold goes down as $D^U$ increases.
{\bf B},{\bf E}, Scatter plots of all words with $N_w>5$ in both windows (12,883 words for comp.os.linux.misc,
12,237 words for rec.music.hip-hop).
Values on $y$-axis: log-frequency change $\Delta \log_{10} f \equiv \log_{10}f(t_2) -
\log_{10} f(t_1)$. Red lines: running median, 10th percentile, and 90th percentile. 
Words with rising frequency appear above and words with falling frequency appear below $\Delta \log_{10}f = 0$.
Examples of words with large frequency changes are highlighted. 
The probability of frequency decay is greater for words with low $D^U$.
{\bf C},{\bf F},  Summary of the dominant pattern in panels {\bf B},{\bf E} over all non-overlapping windows with $t_1$ ranging 
from the beginning of the group 
to $2006$, and $t_2=t_1+2$.
Median values of $\Delta \log_{10} f$ at $D^U=0.4$ and $D^U=1$ are shown for each pair of windows.}
\label{figure3}
\end{center}
\end{figure*}

We now examine $D^U$ as a predictor of frequency change for words over two-year periods.  
We first note that $D^U$ is strongly related to the likelihood that
a word with $N_w > 5$ in a window $t_1$ falls below this threshold in a window $t_2$ taken two
years later. This is illustrated for both Usenet groups in Figure~\ref{figure3}AD, where $t_1$ and 
$t_2$ mark the centers of the half-year windows.  The finding is
so statistically robust that it is reproduced for every choice $t_1$ and $t_2 = t_1 + 2\mbox{
}\text{years}$, in both groups. The same pattern is also mirrored in the frequency changes of words
that are above the $N_w > 5$ threshold at both  $t_1$ and $t_2$. Within this group of words
in the selected window of comp.os.linux.misc, $D^U$ is a strong predictor of whether the word
rose or fell in frequency (Figure~\ref{figure3}B). In the selected window of rec.music.hip-hop, $D^U$ 
is likewise a strong predictor of the changes in word frequencies  (Figure 3E).
The consistency of this pattern
over all windows may be seen by comparing $\Delta \log_{10}f$ for words with $D^U = 0.4$ and with
$D^U = 1.0$, values that span the well-populated portion of the range in $D^U$. Words with the
former value tend to decline in frequency ($\Delta \log_{10}f$ is negative), while words with
the latter value tend to maintain or increase their frequencies ($\Delta \log_{10}f$ is near zero or positive).   
There is no $t_1, t_2$ pair for
either dataset in which the effect is reversed (Figure~\ref{figure3}CF).

\begin{figure*}[t!] 
\includegraphics[width=1.0\columnwidth]{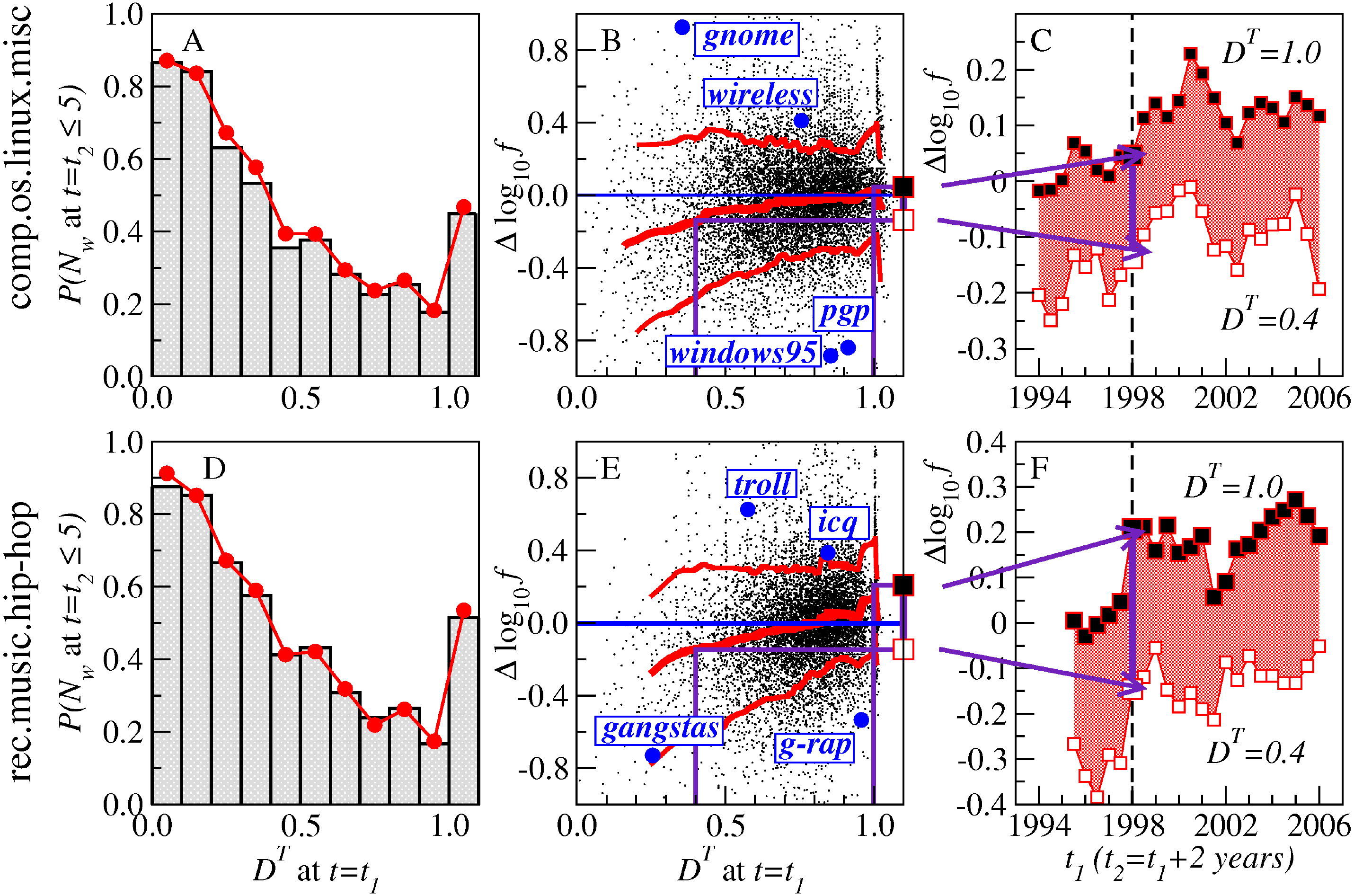}
\caption{ {\bf Dissemination across threads $D^T$ as a predictor of falling below threshold and of frequency decay.}
This figure is the $D^T$-counterpart of Figure~\ref{figure3}.}
\label{figure4}
\end{figure*}

\begin{figure*}[t!] 
\includegraphics[width=1.0\columnwidth]{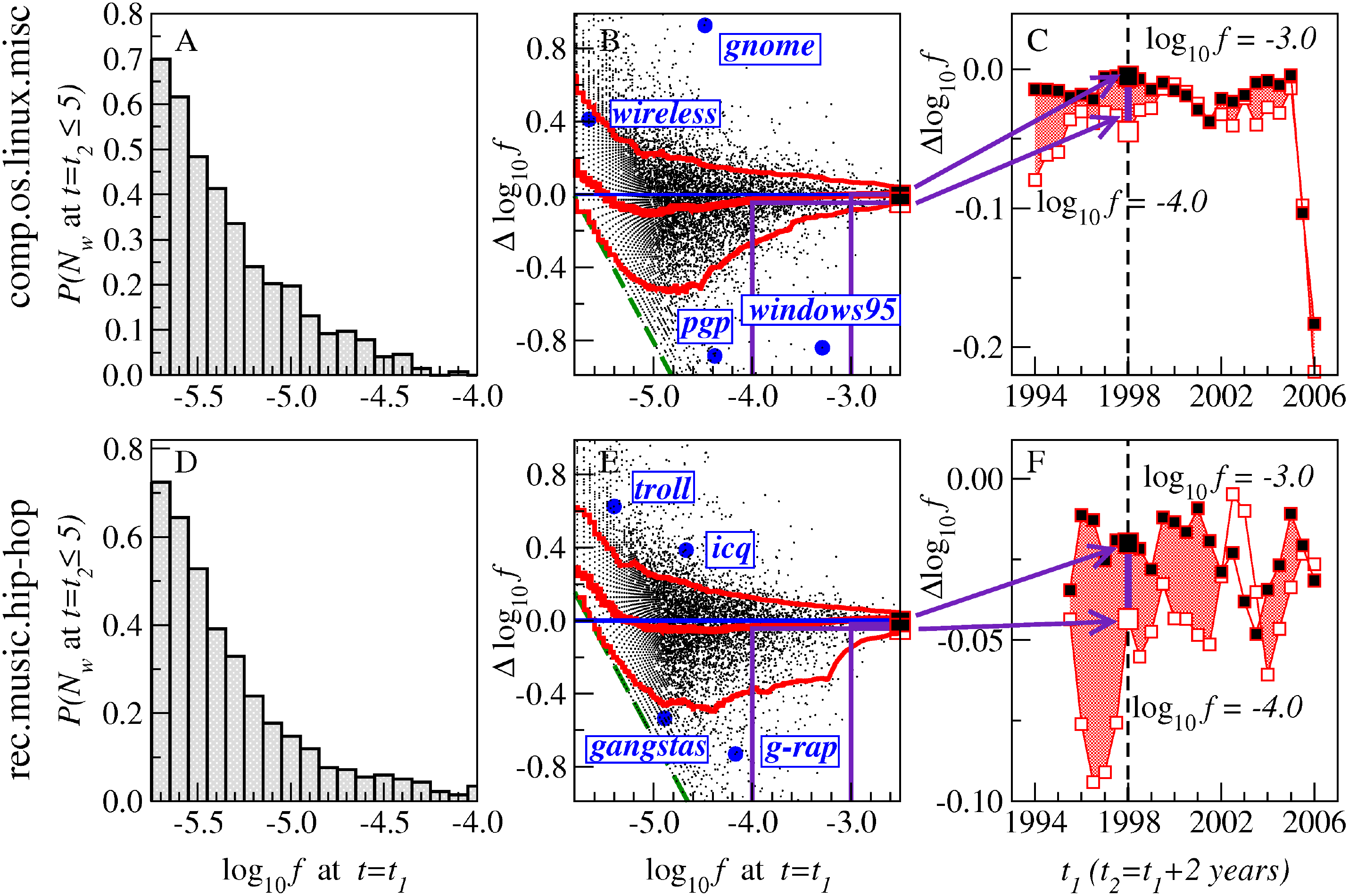}
\caption{ {\bf Frequency~$f$ as a predictor of falling below threshold and of frequency decay.}
This figure is the $f$-counterpart of Figure~\ref{figure3}. The dashed  green lines in panels {\bf
B},{\bf E}
indicate the minimum possible $\Delta \log_{10}f$ for a given~$\log_{10}f(t_1)$, due to the
threshold~$N_w> 5$ imposed at~$t_2$. The analysis in Table~1  includes
only the range $\log_{10}f_{\min} < \log_{10}f< \log_{10} f_{\max}$, where $f_{\min}$ and $f_{\max}$ are the limits of the range considered.  
The range is truncated at $\log_{10} f_{\max}=-2.52$ because, for words above this frequency,
$N_w$ is so large compared to the number of users or threads that $D$ is not informative.
The range is truncated at $\log_{10}f_{\min}=-4.61$ for comp.os.linux.misc ($\log_{10}f_{\min}=-4.52$ for rec.music.hip-hop) because below these cutoffs
the exclusion of words falling under the threshold (i.e., $N_w \le 5$) introduces artifacts in
the relationship to~$\Delta \log_{10} f$ (c.f. the relationship of the dashed green lines to the
10th percentile line). Specifically, $f_{\min}$
was chosen 
for each dataset 
so that the percentage of words falling below the threshold at  $t_2$ would be less
than~$5\%$ of the words with $\log_{10}f_{\min} < \log_{10}f< \log_{10} f_{\max}$.
}
\label{figure5}
\end{figure*}

This far, our analysis has focused on $D^U$. In sociolinguistic parlance, we have considered the
``indexicality'' of words, that is the extent to which words are associated with individuals or
types of people. Now, let us  also consider $D^T$, our measure of ``topicality'' (dissemination across topics).
As shown in Figure~\ref{figure2}CD and in Figure~\ref{figure4},
the results just described for~$D^U$ also hold for~$D^T$.
  The connection between $D^T$ and frequency change agrees with Ref.~\cite{Chesley}'s study of foreign borrowings in
news articles.
What is the relative importance of
these factors in predicting frequency change?
As Table~\ref{tab.factors} shows, $D^U$ is more
important than $D^T$. Moreover, both are more important than  $\log_{10} f$, whose importance is
comparatively slight, as shown in Figure~\ref{figure5}.

Words change over time not just in their frequency, but also in their
dissemination.  A signal aspect of changes in $D^{U,T}$ is a strong negative
correlation with frequency change ($\Delta \log_{10}f$). For comp.os.linux.misc, the correlations
of $\Delta \log_{10}f$ with $\Delta D^U$ and $\Delta D^T$ are $-0.54$ and
$-0.40$, respectively; for rec.music.hip-hop, $-0.55$ and $-0.39$,
respectively. These negative correlations can be understood by comparing
two scenarios. In one scenario, a word rises in frequency because it becomes
more widely used; it is used by more individuals and/or in the
discussion of more topics. In this scenario, the increase in frequency
is accompanied by steady or increasing values of the dissemination measures $D^{U,T}$. In
a contrasting scenario, a word rises in frequency without a concomitant
increase in the number of users and/or topics, because it is used more
repetitively by the same few people and/or in discussing the same topics. In
this scenario, the increase in frequency
is accompanied by decreasing values of $D^{U,T}$, because the use of the word becomes more and more
concentrated in comparison to what the random baseline would predict. In this
case, 
it follows from Figure~\ref{figure3} that 
the resulting low $D^{U,T}$ puts the word at risk of declining in frequency
thereafter. 
Just as a population that explodes in a narrow ecological niche may
well crash later, it appears that repetitive communications 
are more discounted than emulated by others. 
This picture broadly resembles 
recent observations about buzzwords in the blogosphere,
which are reported in Ref.~\cite{neuman} to exhibit
great fluctuations
in their frequencies, as well as an apparent association between a fast rise 
and subsequent obsolescence. 
The fact that the
correlations of frequency change ($\Delta \log_{10}f$) with dissemination
  change ($\Delta D^U$ and $\Delta D^T$) are
strongly negative means that the second scenario is the dominant one in our
datasets. Overall, fluctuations in frequency
driven by variability in user behavior
and topic dominate the statistical behavior, with the result
that patterns similar to those in Figures~\ref{figure3} and~\ref{figure4} are also observed by making the
same calculations in the reversed time direction (that is, by relating $D^{U,T}$ at
$t=t_2$ to $-\Delta \log_{10} f$). These large, short-term fluctuations add an
important new dimension to the study of the long-term dynamics of language, as
any novel expression must survive in the short term to survive in the long term.

\subsubsection*{Case study: Rising slang and product words}
A new word
must establish itself in a niche to survive in the language. The
survival rate of lexical innovations is not known, but any successful innovation must have
overcome short-term fluctuations in $f$ that risked driving it to an early extinction. 
We now
present a case study of successful innovations. 
First we identify all words that were not used during the first years of the group,
and that were consistently used for at least some years thereafter (for precise thresholds, see Supporting Information S1, Text~S2). From this collection of rising words, we selected 
two sets of words for each group.
The first set is
designated as P-words because they refer to products (such as  {\it gnome}, a desktop environment introduced in 1998)
and public figures (such as {\it eminem}, a rapper popular from the late 1990's). Exogenous factors contribute strongly to their
use. 
The second set, designated as S-words, exemplifies slang words and other novel vernacular language. 
These novel words were selected with the
aid of on-line dictionaries of Internet and Usenet terms (see Supporting Information S1, Text~S2).  
We consider the dynamics of these words to be more dominated by factors endogenous to the 
linguistic systems and social networks of the Usenet groups. Although many of the S-words
may have been learned from people outside of a Usenet group, such as celebrities seen on television, 
the group itself is the locus of the the social values and
conventions that lead to some celebrities being imitated and others ignored.
Paired lists of P-words and S-words were frequency matched to the extent possible. The  words and their statistics are 
listed in Supporting Information S1, Tables~S1-S4.

\begin{figure*}[!ht] 
\begin{center} 
\includegraphics[width=0.9\columnwidth]{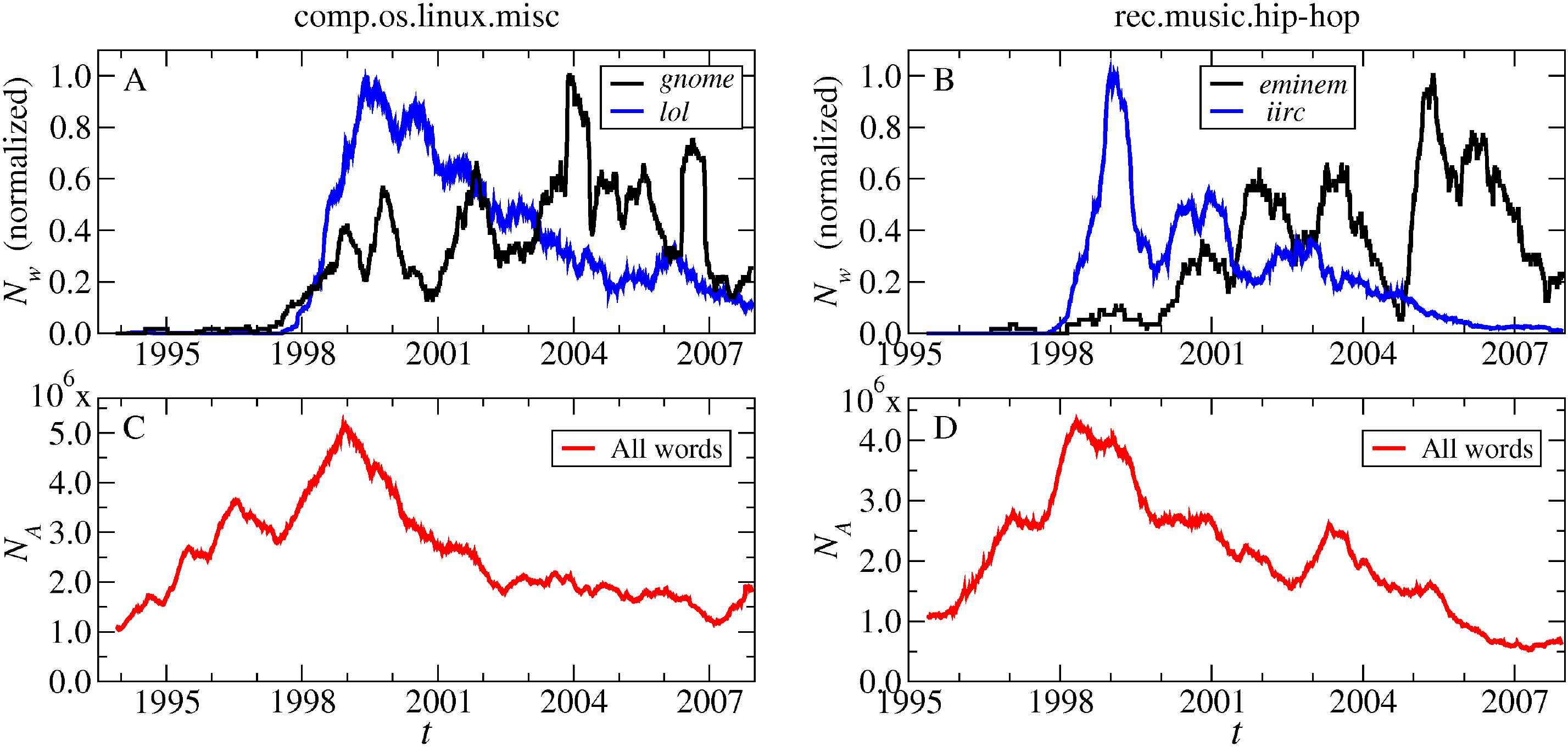}
\caption{\baselineskip 14pt
{\bf Dynamical behavior of P- and S-words in time.}
{\bf A},{\bf B}, Number of occurrences of example P- and S-words as a function of the center $t$ of each half-year window. 
Example words: P-word {\it gnome}, a software product; S-word {\it lol} (``laughing out loud"); P-word {\it eminem}, a rapper; 
S-word {\it iirc} (``if I recall correctly"). 
The curves are normalized by the maximum number of occurrences per window reached over all windows: 
$1,360$ for {\it gnome} and $115$ for {\it lol} ({\bf A}); 
$2,510$ for {\it eminem} and $56$ for {\it iirc} ({\bf B}). 
{{\bf C},{\bf D}, Total number~$N_A$ of all words in each half-year window centered at~$t$.}
}
\label{figure6}
\end{center}
\end{figure*}

\begin{figure*}[t!] 
\includegraphics[width=1\columnwidth]{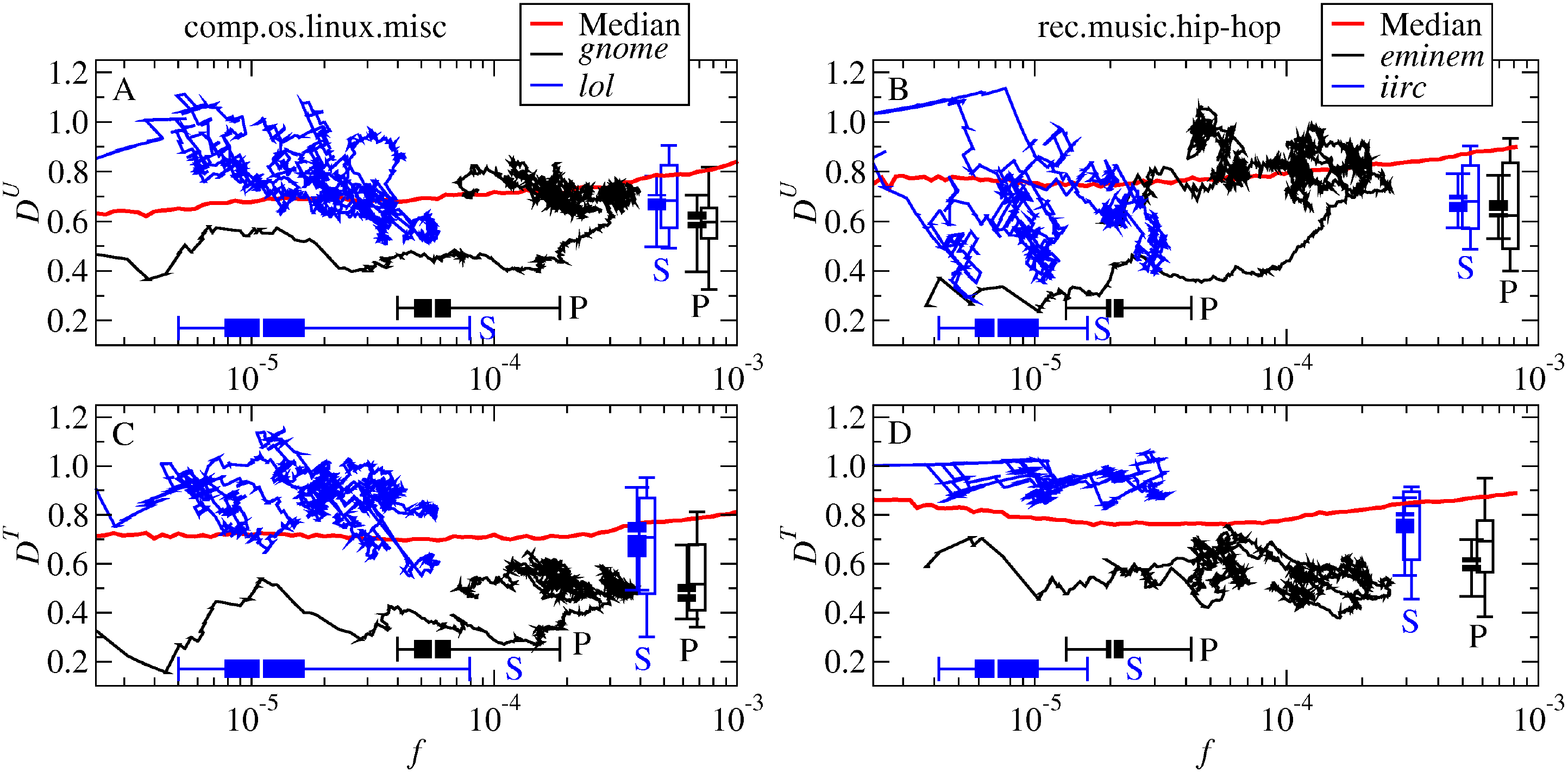}
\caption{ {\bf Dynamical behavior of P- and S-words in frequency and dissemination.} 
{\bf A},{\bf B}, Relationship of $D^U$ to frequency. 
Black and blue curves: evolution of example P-words and S-words over time.
Red line: median over all words, as in Figure~\ref{figure2}.
Boxplots: distribution of the  mean frequency~$f$ (solid, horizontal), mean dissemination~$ D^U$ (solid, vertical),
and mean dissemination~$ D^U$ in the rising period (open, vertical) for all P- and S-words (Supporting Information S1, Tables~S1-S4).
The mean is calculated over all words with $N_w>5$ within the corresponding window.
{{\bf C},{\bf D}, The $D^T$-counterpart of panels {\bf A},{\bf B}.}}
\label{figure7}
\end{figure*}

Figure~\ref{figure6} compares the dynamics of example P-words and S-words. Temporal fluctuations in the total activity of the
group  (Figure~\ref{figure6}CD) provide a backdrop for considering the different fluctuations in
the number of occurrences of some typical P-words and  S-words (Figure~\ref{figure6}AB).
Our Usenet database also allows us to go beyond the frequency dynamics of words over
time, as explored in Ref.~\cite{Michel}'s recent study of words in books,
and look at the roles of topics and individuals in determining this dynamics.
In Figure~\ref{figure7}, we show the behavior of the words in a
frequency-$D^{U,T}$ space. As indicated by the horizontal boxplots, the P-words and S-words are located
in the frequency region below $\log_{10} f = -3.5$, in which the frequency is not correlated with
$D^{U,T}$. Trajectories over time for two example words are superimposed, beginning when the words first
reach $N_w > 5$. In contrast to the example S-words, the example P-words begin with very low
$D^U$ values, and rise greatly in frequency before becoming widely disseminated. The vertical
boxplots show that P-words have overall lower $D^{U,T}$ than S-words (though
both fall below the median of all words). 
The contrast in $D^{U,T}$ over the entire period is replicated if we consider just the early rising period of
each of the words in both groups (see the aggregated statistics displayed in Figure~\ref{figure7}, and further
details in Supporting Information S1, Tables~S1-S4).

Significant clumping in $D^U$ is expected for S-words, because choices of vernacular language
such as {\it lol} ({\it laughing out loud}) and {\it prolly} ({\it probably}) reflect the
individual's construction of social identity~\cite{milroy, eckert}.  
How can we construe the finding that P-words are even more clumped in $D^U$ than the
S-words are? {Recalling that all of the words in the case study were preselected 
to exemplify rising trends,
it seems possible that the highly clumped P-words reflect the distinctive information access of their users.
For example, {\it gnome}, which has a $D^U$ value of $0.46$ in its
early rising period, refers to a graphical desktop environment that was originally created by two Mexican programmers,
Miguel de Icaza and Federico Mena. By discussing their experience with this interface, 
its early adopters bring information to the comp.os.linux.misc group that other users do not
yet have. In short, by contributing posts about experiences and activities external to the Usenet group, a
small number of users 
can be the vehicle for exogenous factors to come to influence the vocabulary of the group more generally.}

The low $D^U$  of the P-words and S-words would tend to predict a decline in frequency (see above),
but instead the frequencies of these particular words rose. For P-words, the rise is
 driven by events external to the Usenet community. For example, the P-word {\it ssh} (from
comp.os.linux.misc) refers to the secure shell network protocol. The invention of ssh allowed
people to carry out remote file transfers without compromising sensitive information such as
passwords. The immediate adoption of this technological improvement is clearly one reason for the
rise in use of the word {\it ssh}. In rec.music.hip-hop, the use of the P-words {\it bush, saddam, and
iraq} reflects discussion about the war in Iraq. Both the war, and the political events leading up
to it, took place outside of the Usenet community. 
{In Figure~\ref{figure6}B, the 2005 rise in the frequency of {\it eminem} reflects heavy
media coverage of his possible retirement}.
The use of the P-words also reflects endogenous
factors to some extent. The fact that {\it bush, saddam, and iraq} met the inclusion criteria in
rec.music.hip-hop, but not in comp.os.linux.misc, suggests that a shared interest in politics is more
important within the Usenet hip-hop community than in the Usenet linux community. 

However, for the S-words, we consider that the endogenous factors
were even more important. For these words, there are alternative ways
of referring to the same general concept. In both groups, {\it lol}
competes with {\it rofl} (rolling on the floor laughing),  {\it
ha-ha}, and other expressions. In rec.music.hip-hop, {\it addy}
competes with {\it address}.
In comp.os.linux.misc,  {\it y2k} competes with {\it year 2000}, and
{\it boxen} (as a plural of {\it box}, generalizing the jocular
plural of {\it Vaxen} for the {\it Vax} brand of computers) competes
with {\it boxes, servers, computers}, etc.  The choice of one such
word over an alternative expression with the same referent reflects
the social value associated with the word, which is a non-referential
component of its meaning. By their nature, slang words stand out from
other words through being used to ``establish or reinforce social
identity or cohesiveness within a group, or with a trend or fashion
in society at large"~\cite{eble}. In African-American Vernacular
English (the original language of hip-hop), the transitory slang
expressions of various subgroups of speakers, such as teenagers and
musicians, serves to differentiate them within a larger African-American 
community sharing a rather stable lexicon and grammar~\cite{smitherman}. 
Reference \cite{Crystal} suggests that on-line groups
are especially likely to use jargon and slang as a means of
constructing and affirming group solidarity, since the group has no
identity outside of its on-line communications. But the use of some
S-words also reflects exogenous factors to some extent, which may 
help explain their success despite the relatively low dissemination. 
The invention
of cell-phone texting probably contributed to the availability of
acronyms as slang expressions,  the rise of server farms probably
contributed to the need for a way to refer to  computers as fungible
units, and the linguistic influence of a particular  rapper might
have increased after a successful performance.  However, these
factors seem weaker than for the P-words, because they do not appear
to dictate the particular choice of word out of all the alternatives.
Related cases of social dynamics  for which a combination of
exogenous and endogenous factors has been considered 
include music
downloads~\cite{watts} and popularity patterns for  YouTube videos and for
stories on the news  portal Digg ~\cite{lerman,szabo}.

By having the lowest overall distribution of $D^U$ values, the P-words
contrast with all other rising words, including both the S-words and
typical words {whose frequencies increased 
(as exemplified in
Figure~\ref{figure3}BE by data points in the upper-right quadrant of each panel).}
This suggests that exogenous forcing is more efficient than 
other kinds
of forcing. The fact that S-words had higher $D^U$ values overall
than the P-words did, with no S-word rising from as low a $D^U$ value as the 
lowest P-words, makes the S-words appear more similar to words in general.
In the absence of strong forcing by external events, the social dynamics within the group
dominates the word dynamics, with reinforcement by peers providing a natural 
mechanism for the words to rise.  The results {support}  our understanding of $D^U$ as a 
determinant of frequency change; high $D^U$ values provide an index of the
fact that relatively many different users provide examples {of use of a specific
  word}
that others may imitate.
The $D^U$ values for S-words are somewhat low compared to the distribution for all words. 
We can speculate about the mechanisms for this outcome. Exogenous factors in the use of
S-words, mentioned just above, may play a greater role 
than is typical for words in general. 
Moreover, the force and emotions associated with the social value of the S-words may provide
an additional factor 
driving 
the dynamics. 

Most of our principal observations about the dissemination across users
($D^U$) of P-words and S-words are also true for the dissemination  of
the same words across topics~($D^T)$, as shown by comparing Figure~\ref{figure7}AB to Figure~\ref{figure7}CD. 
Given that the 
measures~$D^U$ and~$D^T$ both quantify the relative extent of the word niche, these detailed 
parallels in the behavior of the two measures raise the question of how many dimensions
we are really dealing with. 
Since people form social groupings around shared interests~\cite{fortunato, schifanella}, 
and choose words that express solidarity with these same groupings, 
do the two dimensions of indexicality and topicality reduce
to just one underlying dimension? 
Or are the two
dimensions separable, even if related through complex interactions? We take up these questions
rigorously in the next section.

\subsubsection*{Factoring the relative contributions of individuals and topics}
We have shown that most words, including both highly indexical words such as slang words and highly
topical words such as products, are significantly concentrated in both $D^U$ and $D^T$. We have
sketched some reasons for these dimensions to be positively correlated. How can we rigorously
evaluate their separability and relative importance? To address this issue, we  consider new
measures that effectively factor indexicality and topicality as contributors to $D^{U,T}$, and
we standardize the datasets to eliminate distributional artifacts.

We first introduce $\hat{D}^U$ as a modification of $D^U$ in which $\tilde{U}$ in
Eq.~(\ref{eq.dissemination}) is calculated from a baseline model that shuffles the words only
within threads, rather than across all users and all threads. Analogously, we
introduce $\hat{D}^T$ as a modification of $D^T$ in which $\tilde{T}$ in
Eq.~(\ref{eq.disseminationT}) is calculated from a baseline model that shuffles the words only
within posts of the same user. These new quantities provide a direct measure of the extent to which
individuals and topics contribute to the concentration of words observed above. While $D^U$ reveals
whether the word is clumped or over-disseminated by comparing the actual dissemination with that
obtained by ``erasing" all the structure, $\hat{D}^U$ maintains the structure of the threads and
considers randomization of words across users within them. If $\hat{D}^U$ is significantly closer
to $1$ than $D^U$ is, then  topics must strongly influence the individuals' choice of words.
Analogously, the role of individuals can be confirmed by comparing the extent to which $\hat{D}^T$
is closer to $1$ than $D^T$ is.

To ensure that users and threads serve as comparable proxies of individuals and topics,
we randomly trim the datasets to eliminate the differences in their distributions that are visible in Supporting Information S1, Figure~S1. For each window, the
trimming scheme standardizes the user contribution per thread and the size of all posts, matches
the number of users and threads, and approximately matches the distribution of posts per user and
per thread (see Supporting Information S1, Text~S3 and Figure~S2). The trimmed comp.os.linux.misc (rec.music.hip-hop) dataset remains
large enough for our statistical analysis, with an average of $4,593$ ($1,503$) posts
and $2,383$ ($585$) users and threads per half-year window, and an overall average of $77.6$
($51.2$) words per post. 

The exact distributions of values of $D^U$ and $D^T$ change with the trimming. Trimming generally
increases $D^U$ and $D^T$ for the words that survive, but the trends and all conclusions from
previous sections still stand. 
For example, the overall median $D^U$ changes from $0.71$ to $0.87$,
and the overall median $D^T$ changes from $0.73$ to $0.89$, for the comp.os.linux.misc group. The
relative differences in both groups remain essentially unchanged, which means that 
the measures $D^{U,T}$ provide meaningful comparisons even when the distributions are not streamlined. However,
the trimmed set offers the advantage of providing exact and non-artifactual information about the
correlations between the measures.

Table~\ref{tab.corr} displays the important correlations amongst the original and modified
measures. The correlation between $D^U$ and $D^T$ is positive, confirming the expectation that
indexicality and topicality are related. But it is far less than $1$, suggesting that $D^U$ and
$D^T$ contribute substantially different information. The measures $D^U$ and $\hat{D}^U$, as well as
$D^T$ and $\hat{D}^T$ are positively correlated, as expected because these are related measures by
definition.  Finally, the negative correlation between $\hat{D}^U$ and $\hat{D}^T$ is a
confirmation that these quantities partially factor $D^U$ and $D^T$ and hence provide the
information they are designed to provide. { Notice that this negative correlation is
  possible, despite the positive correlation of the other pairs of variables, because the
  positive correlations are not all close to one.}

\begin{figure*}[!ht] 
\begin{center} 
\includegraphics[width=0.8\columnwidth]{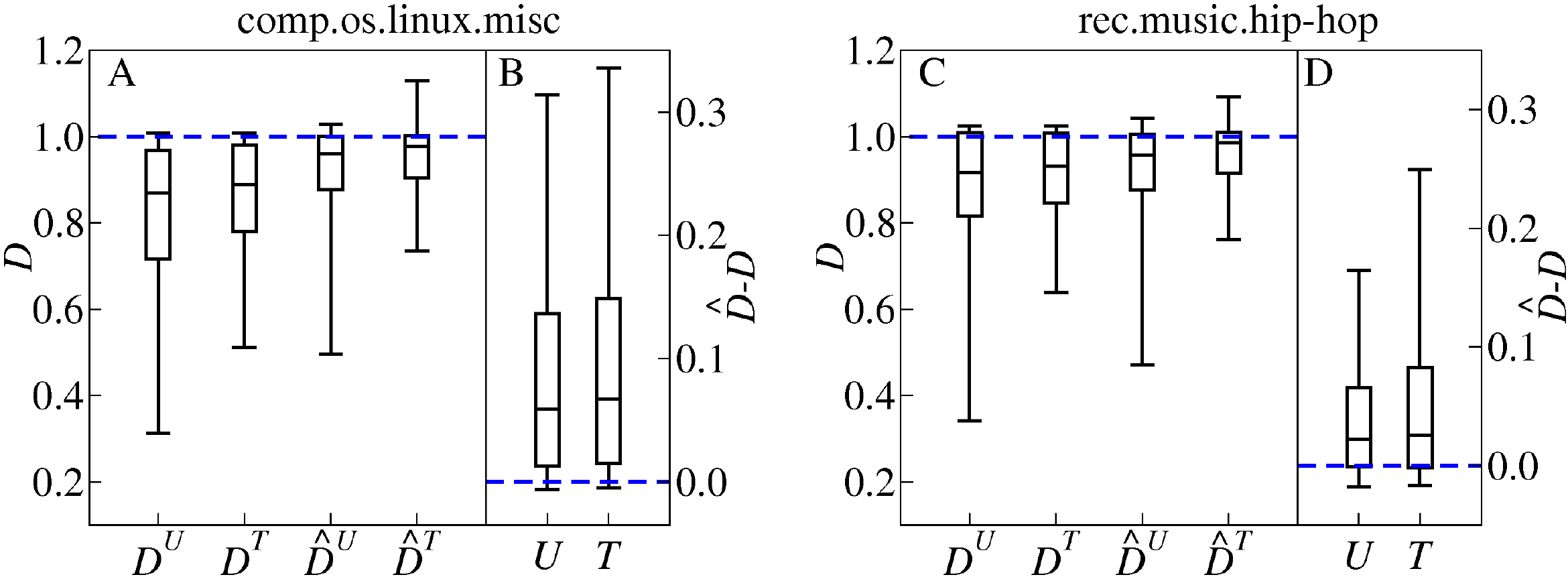}
\caption{\baselineskip 14pt
{\bf Summary statistics of the dissemination measures.}
{\bf A},{\bf C}, The box-and-whisker plots indicate the median, the quartiles, and the octiles for $D^{U,T}$ and $\hat{D}^{U,T}$ over the collection of all non-overlapping windows of the trimmed datasets. {\bf B},{\bf D}, Corresponding statistics for $\hat{D}^{U,T}-D^{U,T}$ estimated from individual words. The statistics includes all words with $N_w>5$ within the corresponding windows, 
with occurrences in different windows being counted independently. 
}
\label{figure8}
\end{center}
\end{figure*}

We now use the trimmed datasets and modified measures to further test the relative importance of
indexicality and topicality. As shown in Figure~\ref{figure8}AC, $\hat{D}^U$ and $\hat{D}^T$  are
statistically larger than $D^U$ and $D^T$, respectively, but they remain smaller than $1$. This confirms that
most words are clumped with respect to both users and threads. Overall, $D^U$ is smaller
than $D^T$, indicating that words are generally more concentrated with respect to users than to
threads. 
This observation is rigorously confirmed by the fact that $\hat{D}^U$ is smaller than
$\hat{D}^T$ to a comparable extent as $D^U$ is smaller than
$D^T$. {Figure~\ref{figure8}BD shows that also for individual words, $\hat{D}^U$ and
  $\hat{D}^T$  are typically larger than $D^U$ and $D^T$, respectively. Furthermore,}
  {we can elucidate the effect of threads on users by considering the magnitude of the
  difference $\hat{D}^U - D^U$, and similarly, the effect of users on threads by considering
  $\hat{D}^T - D^T$. These comparisons reveal that} {the effect of threads 
on users is statistically smaller than the effect of users on threads, both in the aggregate 
(Figure~\ref{figure8}AC) and for individual words (Figure~\ref{figure8}BD).}

The most striking effect shown in Figure~\ref{figure8}AC is the large number of words with small
$D^U$ in comparison to $D^T$. After trimming, over all windows, the comp.os.linux.misc
(rec.music.hip-hop) dataset has $5,356$ ($1,808$) words with $D^U<0.4$, versus $1,657$ ($337$)
words with $D^T<0.4$. The list of words with $D^U<0.4$ but $D^T>0.4$ includes both very common
words and highly topical words. In comp.os.linux.misc, example words include {\it imagination},
{\it coffee},  {\it angst-ridden}, and {\it saukrates} (a rapper); in rec.music.hip-hop, examples
include {\it regards}, {\it baptized} and {\it tauri} (a Hungarian Warcraft server). It is
interesting that such words are even more distinctive to individuals than to topics. A contributing
factor to this clumpiness is the use of formulaic expressions. Such expressions, which are found in
signature blocks, as well as in other conventionalized communications like greetings and
insults, often have quite idiosyncratic lexical choices.

Altogether, we have strong evidence that the lexical make-up of the threads is strongly determined
by the individual users. This speaks against the possibility that the topic dictates the 
vocabulary, and equally against the possibility that mutual imitation causes strong convergence in
lexical choices as people interact in the discussion. This is a striking result. It contrasts with
the major thrust of research on modeling the evolution of lexical systems, which is to explain
convergence in the community~\cite{Steel, komarova}. This suggests that individuals may be more autonomous 
in their choices of words than
in a wide range of other behaviors, from yawning and gait~\cite{dijksterhuis} to 
complex conscious decisions like the decision to purchase a product or to vote~\cite{Nickerson}.
Given that individuals use different words to talk about the same topic,
that word concentration over users is more extreme than over threads, and that $D^U$ is
the strongest predictor of {frequency change}, 
the heterogeneity of people emerges as the single
strongest factor in lexical diversity, both at any particular time and over time.

\section*{Discussion}

We have introduced two new quantities, $D^U$ and $D^T$, as measures of the dissemination of words
across individuals and topics, and used them to characterize the vocabulary of two online
discussion groups over a period of more than a decade. We found that almost all words are
concentrated with respect to both individuals and topics, and that at short-term (two-year) time
scales, the word's concentration in the space of users and topics, as revealed by  $D^{U,T}$,
 is a strong determinant of word fate. $D^U$ and $D^T$
are separable components, and both trump word frequency. However, $D^U$ trumps
$D^T$. 

Word frequencies over time reflect a replicator dynamic, that is, a dynamic in which the words are
reproduced by being copied through imitation~\cite{Steel, pop1,komarova, kirby}.  
Including both learning and use, this dynamic reflects an interaction of 
social and cognitive factors~\cite{hruschka}. Word learning is facilitated by variety in the context of
use~\cite{blythe}, and rates of word use are in turn subject to great fluctuations over time, as a
reflex of shifting user behavior and shifting topics. For a lexical innovation to survive in the
language, it must avoid an absorbing boundary near $f=0$, at which it is used so rarely
that no one can learn it. Our investigation of the relationship between frequency
  change 
and dissemination change
shows that a key to success beyond short-term fluctuations is increasing
  frequency~($f$) hand-in-hand with increasing dissemination ($D^{U,T}$). The success of the P-words in our case study can be understood by considering that
exogenous forcing by external events allowed them to overcome the handicap of low
dissemination
values.  
S-words, selected to exemplify more endogenous dynamics, behaved more like words in general
by displaying higher dissemination
values when rising. 

Word frequency affects word fate at historical time scales when
different forms compete to express the same meaning~\cite{Pagel,
Lieberman, Michel}.  Why did frequency not prove to be important in the
dynamics of the whole vocabulary, as studied here? The language
system has strong functional pressures for words to be distinct from
each other, in both form and meaning~\cite{davis, Steel, komarova,
kirby,baronchelli}. Although dictionaries use words to explain the
meanings of other words, and thesauri group together words with
related meanings, true synonymy is very rare~\cite{clark, wexler}.
For words which might seem to be synonyms, such as {\it soda} vs.\
{\it pop}, or {\it yes} vs.\ {\it yup}, there is normally a difference
in dialect, formality, or other contextual factors governing the use
of the word. Because almost every word is learned with a distinctive
meaning (or set of meanings), and replication has low error rates, it
follows that most words do not have a direct competitor for exactly
the same meaning and contexts of use. If an active competition
between two forms develops historically, then both can survive if
they develop distinctive roles within the space of the lexical,
syntactic, and pragmatic components of the linguistic system. For
example, the English future auxiliary {\it gonna} is a new competitor
for the older future {\it will}, but both survive because {\it gonna}
is preferentially used in some constructions (such as questions),
whereas {\it will} is preferentially used in others (such as the main
clauses of conditionals)~\cite{torres}. Reference~\cite{torres} indeed
uses the term {\it niche} to characterize these distinctive
components in the usage of different future expressions, suggesting
that differentiated niches are critical to their ongoing use in the
language. These results complement those presented here by analyzing
dimensions of the word niche that are 
internal to the linguistic system. The picture presents
strong parallels to the exclusion principle in evolutionary biology,
which states that occupying distinct niches protects species from
competition~\cite{hardin}. Similar reasoning can also be applied
to explore the competition between entire languages. In a  model of
language competition that assumes the speakers to be monolingual,
distinct languages are similarly predicted to survive only if they
are spoken by distinct, partially unmixed populations~\cite{abrams}.
This prediction is attenuated if bilingualism in itself has high
value or status as a human capability~\cite{sole}, permitting
bilinguals to occupy a social position that is not available to
monolinguals.

Diversity therefore depends on the
diversity and viability of the individual niches. For biological species, the
size of the geographical range and the species duration are
correlated~\cite{foote, jablonski}. In studies of the lexicon, the individual
words assume the role of species, and we have shown that 
the relative extent of the word niche is associated with 
the likelihood of a favorable or unfavorable fate.
But we have also shown that the relative extent of the word niche does not provide the whole story 
about viability. In population biology, exogenous events such as asteroid impacts can overcome
the general statistical trends associated with dissemination. The same thing is true here, where exogenous 
events such as inventions and wars can overcome general statistical trends associated with the dissemination
of words.
This generalization is further illustrated
by the  recent finding that censorship can induce large and distinctive deviations from typical
frequency 
trajectories for the names of people~\cite{Michel}.

We found that $D^U$ and $D^T$ are positively correlated, but still provide distinct information.  A
positive correlation is expected because individuals have characteristic interests. Further
mechanisms contributing towards this correlation result from the participation of individuals in
social and geographical structures.  For example, these can cause clumping in product use, as shown by 
profiling the Internet for software products~\cite{trestian}, which entails clumping of the words
used to discuss those products. 
Structures in the
social network can even contribute directly to product adoption, because the usefulness of many
products (such as high-tech innovations) can depend on the number of neighbors  who already use the
product~\cite{watts, rogers}.  
These same mechanisms pertain to other words, insofar as concepts and opinions resemble
products. 

We suggest, however, that other mechanisms limit the correlation between~$D^U$ and~$D^T$, and explain the
striking degree to which individuals were found to use different words in discussing the same
topic. 
The variety in human social identities is thought to provide an impetus for innovation in
modes of expression, as discussed in classic works of sociolinguistics
\cite{milroy, eckert, labov}.  Because people tend to associate with people like
themselves, the variety in social identities can also give rise to clusters within social
networks~\cite{wasserman}, and these
clusters can in turn
hinder lexical convergence~\cite{labov, lu, hruschka}.
The fundamental principles of discourse call for one to strike a
balance between anchoring contributions in what the listener already knows, and providing
novel and relevant information~\cite{Grice}. Online discourse can be viewed as a collective
exploration of the conceptual world~\cite{cattuto2}. 
It follows from this study that the most engaging and fruitful discourse is discourse in
which people cooperate in differentiating themselves and what they say.

\section*{Methods} \label{ssec.methods}

\noindent
{\it Datasets.}\  Usenet group archives are available at http://groups.google.com. The smallest unit of text 
is the {\em post}. Each post is attributed to a {\em user} and belongs to a {\em thread} (as
defined by an initial post and all replies to it). We focus on two Usenet groups from their first
post through $2008\mbox{-}03\mbox{-}31$: (i) comp.os.linux.misc, which concerns Linux operating systems, includes
$128,903$ users and $140,517$ threads beginning $1993\mbox{-}08\mbox{-}12$; (ii) rec.music.hip-hop, which
is devoted to hip-hop music, has $37,779$ users and $94,074$ threads beginning $1995\mbox{-}02\mbox{-}08$.
The activity of users in Usenet groups is bursty~\cite{Altmann} and heterogeneous~\cite{B1}. In
the comp.os.linux.misc group, for example, the average user contributes~$5.4$ posts and remains active for $249.3$ days,
but the most persistent users have more than $1,000$ posts over more than $10$ years. The average
thread has $4.9$ posts and is active for~$4.5$ days, but the longest threads have more than
$1,000$ posts over $3$ years. See Supporting Information S1, Text~S1 for information about preprocessing
of the text, and Figures~S1 and S3 for information about the
fat-tailed distributions that characterize these groups.

\noindent
{\it Baseline model.}\ 
The expected number of users~$\tilde{U}$ in Eq.~(\ref{eq.dissemination})
is calculated by assuming that all  words are randomly 
shuffled, while holding constant the number of users and the number of words per 
user. Let~$N_w$ be the number of occurrences of the word~$w$, $m_i$ be the total number of words 
contributed by user~$i$, and $N_A\equiv\sum_i m_i=\sum_w N_w$. 
The probability that the~$j+1$~th occurrence of~$w$ does not belong to user~$i$ is 
given  by~$(1-m_i/(N_A-j))$. The probability~$\tilde{U}_i$ that user~$i$ used word~$w$ at 
least once is calculated as the complement of the probability
of not using it:
\begin{equation}\label{eq.tildeui}
\tilde{U}_i=1-\prod_{j=0}^{N_w-1}\left(1-\frac{m_i}{N_A-j}\right) \approx 1-e^{-f_w m_i},
\end{equation}
where the approximation is valid for~$m_i/N_A\ll1$ and~$f_w\equiv N_w/N_A\ll1$. 
This corresponds to a Poissonian baseline model with a fixed probability of using~$w$ given by
the observed word frequency~$f_w$.
The error in the approximation is smaller than $0.1\%$ for the 
datasets we consider. This approximation was used in all calculations involving the untrimmed 
datasets, while the exact relation was used for the trimmed datasets.  
An analogous procedure is used for the calculation of the expected number of threads~$\tilde{T}$.

\section*{Acknowledgments}

E.G.A.\ was supported by the Northwestern Institute on Complex Systems 
and a Max Planck Society Otto Hahn Fellowship,
J.B.P.\ by JSMF Grant No.\ 21002061, and 
A.E.M.\ by NSF Grant No.\ DMS-0709212 and a Sloan Research Fellowship.

\section*{Supporting Information}

Supplementary methods, references, tables, and figures are available at:\\
http://www.plosone.org/article/info\%3Adoi\%2F10.1371\%2Fjournal.pone.0019009\#s5

\newpage$\phantom{.}$

\newpage$\phantom{.}$

\newpage$\phantom{.}$

\newpage$\phantom{.}$

\newpage$\phantom{.}$

\newpage$\phantom{.}$

\newpage$\phantom{.}$

\newpage$\phantom{.}$

\newpage$\phantom{.}$

\begin{table}[!ht]
\caption{\bf Relative importance of dissemination across users, dissemination across threads, and frequency in word dynamics.}
\begin{tabular}{c|c|c|c}
\hline
Group & $D^U$  & $D^T$  & $\log_{10}f$  \\
\hline
\hline
comp.os.linux.misc & $9.9\%$& $3.5\%$ & $0.2\%$ \\
\hline
rec.music.hip-hop & $22.0\%$& $5.0\%$ & $0.4\%$\\
\hline
\end{tabular}
\begin{flushleft}
 Relative importance of the three factors as predictors of frequency change ($\Delta \log_{10} f$),
 calculated  using the method of Ref.~\cite{kruskal}.  Importance is
 based on the fraction of the variance of $\Delta \log_{10} f$ explained by each
 factor.
 This method conservatively
 estimates the relative 
 importance of the independent variables in a multiple regression setting.  The data are combined over 
 all window pairs  $t_1, t_2 = t_1 + 2$ considered in Figure~\ref{figure3}. 
 To avoid  artifactual correlations for small and large $f$, the range of words is  restricted in~$f$,  as indicated in the caption of Figure~\ref{figure5}.
\end{flushleft}
\label{tab.factors}
\end{table}

\begin{table}[!ht]
\caption{\bf Correlations between dissemination measures. ~~~~~~~~~~~~~~~~~~~~~~~~~~~~~~~~~~~~~~~~~~~~}
\begin{tabular}{c|c|c|c|c}
\hline
Group & ($\hat{D}^U,D^U$)  & ($\hat{D}^T,D^T$)  & ($D^U,D^T$) &($\hat{D}^U,\hat{D}^T$) \\
\hline
\hline
comp.os.linux.misc & $0.82 \pm 0.07$ & $0.67\pm 0.04$ & $0.54 \pm 0.12$&$-0.30 \pm0.01$ \\
\hline
rec.music.hip-hop & $0.94 \pm 0.02$ & $0.83\pm0.10$ & $0.44 \pm 0.09$& $-0.23 \pm 0.11$ \\
\hline
\end{tabular}
\begin{flushleft}
To obtain the correlations, first we calculate $\hat{D}^U,D^U,\hat{D}^T,D^T$ for all 
  words with $N_w>5$ in the half-year windows of the trimmed datasets. 
The Pearson correlation coefficient, for each pair of variables, is then calculated over all words. The values reported in the table correspond to
the averages $\pm$ standard deviations calculated over all
non-overlapping half-year windows. 
\end{flushleft}
\label{tab.corr}
\end{table}


\begin{thebibliography}{99}

\baselineskip 15pt

\bibitem{onnela} 
Onnela J-P, Saram\"aki J, Hyv\"onen J, Szab\'o G, Lazer D, et al. (2007)
Structure and tie strengths in mobile communication networks. 
Proc Natl Acad Sci USA {\bf 104}: 7332--7336.

\bibitem{Barabasi} 
Gonz\'alez MC, Hidalgo CA, Barab\'asi A-L (2008)
Understanding individual human mobility patterns. 
Nature  {\bf 453}: 779--782.

\bibitem{malmgren} 
Malmgren RD, Stouffer DB, Motter AE, Amaral LAN (2008)
Poissonian explanation for  heavy tails in e-mail communication. 
Proc Natl Acad Sci USA  {\bf 105}: 18153--18158.

\bibitem{seshadri} 
Seshadri M, Machiraju S, Sridharan A, Bolot J, Faloutsos C, et al. (2008)
Mobile call graphs: Beyond power-law and lognormal distributions. 
Proc 14th ACM SIGKDD Intl Conf Knowl Disc Data Min. pp. 596--604.

\bibitem{kuiper} 
Kuiper K  (2006)
Knowledge of language and phrasal vocabulary acquisition. 
Behav Brain Sci {\bf 29}: 291--292.

\bibitem{davis} 
Davis MH, Gaskell MG (2009)  
A complementary systems account of word learning: Neural and behavioral evidence.
Phil Trans R Soc B {\bf 364}: 3773--3800.


\bibitem{munat} 
Munat J (2007) Lexical Creativity, Texts and Contexts. Amsterdam: John Benjamins. 


\bibitem{kleinberg} 
Kleinberg J (2003)  Bursty and hierarchical structure in streams. 
Data Min Knowl Dis {\bf 7}: 373--397.

\bibitem{cattuto} 
Cattuto C, Loreto V, Pietronero L (2007) 
Semiotic dynamics and collaborative tagging.
Proc Natl Acad Sci USA {\bf 104}: 1461--1464.

\bibitem{hotho}
Hotho A, J\"aschke R, Schmitz C, Stumme G (2006) 
Information retrieval in folksonomies: Search and ranking. 
Lec Notes Comput Sc  {\bf 4011}: 411--426.


\bibitem{neuman}
Neuman Y, Nave O, Dolev E (2011) Buzzwords on their way to a tipping-point: 
A view from the blogosphere. Complexity {\bf 16}: 58--68.


\bibitem{Crystal} 
Crystal D (2006) Language and the Internet. Cambridge: Cambridge Univ. Press.

\bibitem{B1} 
Fisher D, Smith MA, Welser HT (2006) 
You are who you talk to: Detecting roles in Usenet Newsgroups. 
Proc 39th Annual Hawaii Intl Conf Syst Sci {\bf 3}: 59.

\bibitem{Pagel} 
Pagel M, Atkinson A, Meade A (2007)
Frequency of word-use predicts rates of lexical evolution throughout Indo-European
history.  Nature {\bf 449}: 717--720.

\bibitem{Lieberman} 
Lieberman E, Michel JB, Jackson J, Tang T, Nowak MA (2007) 
Quantifying the evolutionary dynamics of language.
Nature {\bf 449}: 713--716.

\bibitem{fontanari} 
Fontanari JF, Perlovsky LI (2004)
Solvable null model for the distribution of word frequencies. 
Phys Rev E {\bf 70}: 042901.

\bibitem{Colwell} 
{ Colwell RK, Futuyama DJ (1971) On the measurement of niche breadth and overlap. 
Ecology {\bf 52}: 576-576.}


\bibitem{foote} 
Foote M, Crampton JS, Beu  AG, Cooper RA (2008)
On the bidirectional relationship between geographic range 
and taxonomic duration. Paleobiology {\bf 34}: 421--433. 

\bibitem{jablonski} 
Jablonski D (2005)
Mass extinctions and macroevolution.
Paleobiology {\bf 31}: 192--210.


\bibitem{pop1}
Nettle D (1999)
Is the rate of linguistic change constant?
Lingua {\bf 108}: 119--136.

\bibitem{pop2}
Wichmann S, Stauffer D, Schulze C, Holman EW (2008)
Do language change rates depend on population size?
Adv Complex Syst {\bf 11}: 357--369.

\bibitem{pop3}
Lupyan G, Dale R (2010)
Language structure is partially determined by social structure.
PLoS ONE {\bf 5} (1): e8559.

\bibitem{watts} 
Watts D, Dodds P (2007) 
Influentials, networks, and public opinion formation.
J Consum Res {\bf 34}: 441--458.

\bibitem{salganik} 
Salganik ML, Dodds PS, Watts DJ (2005) 
Experimental study of inequality and unpredictability in an
artificial cultural market.
Science {\bf 311}: 854--856.

\bibitem{eble} 
Eble C (1996) Slang and Sociability: In-group Language among College 
Students. Chapel Hill: The University of North Carolina Press.

\bibitem{smitherman} 
Smitherman G (2000) Black Talk: Words and Phrases from the Hood to Amen Corner. 
New York: Houghton Mifflin. 

\bibitem{Sornette} 
Sornette D (2005) 
Endogenous versus Exogenous Origins of Crises.
In: Albeverio A, Jentsch V, Kantz H, editors.
Extreme Events in Nature and Society. 
Berlin: Springer. pp. 95--119.

\bibitem{Crane} 
Crane R, Sornette D (2008)
Robust dynamic classes revealed by measuring the response function of a social system. 
Proc Natl Acad Sci USA {\bf 105}: 15649--15653.

\bibitem{church95}
Church KW, Gale WA (1995)  
Poisson mixtures.
Nat Lang Eng {\bf 1}: 163--190.

\bibitem{manning}
Manning CD, Sch\"utze H (1999)
Foundations of statistical natural language processing. Cambridge MA: The MIT Press.

\bibitem{serrano}
Serrano MA, Flammini A, Menczer F  (2009)
Modeling statistical properties of written text.
PLoS ONE {\bf 4} (4): e5372. 

\bibitem{Altmann}
Altmann EG,  Pierrehumbert JB, Motter AE (2009) 
Beyond word frequency: Bursts, lulls, and scaling in the temporal distributions of words. 
PLoS ONE {\bf 4} (11): e7678.

\bibitem{Chesley}
Chesley P, Baayen RH (2010) 
Predicting new words from newer words, Lexical borrowings in French. 
Linguistics {\bf 48}: 1343-1374.

\bibitem{Michel}
Michel J-B, Shen Y K, Aiden A P, Veres A, Gray M K, et al. (2011)
Quantitative analysis of culture using millions of digitized books.
Science {\bf 331}:  176-182.

\bibitem{milroy} 
Milroy L (1980)
Language and Social Networks. Oxford: Blackwell.

\bibitem{eckert} 
Eckert P (2000)
Linguistic Variation as Social Practice. Oxford: Blackwell.

\bibitem{lerman} 
Lerman K, Hogg T (2010) 
Using a model of social dynamics to predict popularity of news.
Proc 19th Intl Conf WWW. pp. 621--630.

\bibitem{szabo} 
Szabo G, Huberman BA (2010) 
Predicting the popularity of online content.
Commun ACM {\bf 53}: 80--88.

\bibitem{fortunato} 
Fortunato S, Flammini A, Menczer F, Vespignani A (2006)
Topical interests and the mitigation of search engine bias.
Proc Natl Acad Sci USA  {\bf 103}: 12684--12689.

\bibitem{schifanella} 
Schifanella R, Barrat A, Cattuto C, Markines B, Menczer F (2010)  
Folks in folksonomies: Social link prediction from shared metadata. 
Proc 3rd ACM Intl Conf Web Search Data Min. pp. 271--280. 


\bibitem{Steel} 
Steels L (1997) 
The synthetic modeling of language origins. 
Evol Commun {\bf 1}: 1--34.

\bibitem{komarova}
Komarova NL, Nowak M (2001) 
The evolutionary dynamics of the lexical matrix. 
Bull Math Biol {\bf 63}: 451--484.

\bibitem{dijksterhuis} 
Dijksterhuis A, Bargh JA (2001)
The perception-behavior expressway: Automatic effects of social perception on social behavior.
Adv Exp Soc Psychol {\bf 33}: 1--40.

\bibitem{Nickerson}
Nickerson DW (2008) 
Is voting contagious? Evidence from two field experiments. 
Am Polit Sci Rev {\bf 102}: 49--57.

\bibitem{kirby} 
Kirby S, Cornish H, Smith K (2008)  
Cumulative cultural evolution in the laboratory: An experimental 
approach to the origins of structure in human language.  
Proc Natl Acad Sci USA {\bf 105}: 10681--10686.

\bibitem{hruschka} 
Hruschka DJ, Christiansen MH,  Blythe RA, Croft W, Heggarty P, et al. (2009) 
Building social cognitive models of language change.
Trends Cogn Sci {\bf 13}: 464--469.

\bibitem{blythe} 
Blythe RA, Smith K, Smith ADM (2010) 
Learning times for large lexicons through cross-situational learning.
Cogn Sci {\bf 34}: 620--642.


\bibitem{baronchelli} 
Baronchelli A, Gong T, Puglisi A, Loreto V (2010) 
Modeling the emergence of universality in color naming patterns. 
Proc Natl Acad Sci USA {\bf 107}: 2403--2407. 

\bibitem{clark} 
Clark EV  (1990)
On the pragmatics of contrast.
J Child Lang {\bf 17}: 417-431.

\bibitem{wexler} 
Wexler P, Culicover P (1980)
Formal Principles of Language Acquisition. Cambridge MA: MIT Press.

\bibitem{torres}
Torres Cacoullos R, Walker JA (2010) 
The present of the English future: Grammatical variation and collocations in discourse.
Language {\bf 85}, 321-353.

\bibitem{hardin} 
Hardin G (1960)  
The competitive exclusion principle.
Science {\bf 131}: 1292--1297.

\bibitem{abrams}
Abrams  DM, Strogatz SH (2003) 
Modeling the dynamics of language death. 
Nature {\bf 24}: 900.

\bibitem{sole}
Sol\'e RV, Corominas-Murtra B, Fortuny J (2010)
Diversity, competition, extinction: The ecophysics of language change.
J R Soc Interface {\bf 7}: 1647--1664.


\bibitem{trestian} 
Trestian I, Kuzmanovic A, Ranjan S, Nucci A (2008)
Unconstrained endpoint profiling (Googling the Internet).
Proc ACM SIGCOMM Conf Data Commun. pp. 279-290.


\bibitem{rogers} 
Rogers EM (2003)  
Diffusion of Innovations. New York: Free Press.

\bibitem{labov} 
Labov W (2001) 
Principles of Linguistic Change: Social Factors. Oxford: Blackwell.

\bibitem{wasserman} 
Wasserman S, Faust K (1994)
Social Network Analysis. Cambridge: Cambridge Univ. Press.

\bibitem{lu} 
Lu Q, Korniss G, Szymanski BK (2009) 
The Naming Game in social networks: Community formation and consensus engineering.
J Econ Interact Coord {\bf 4}: 221--235. 

\bibitem{Grice} 
Grice HP (1975) Logic and Conversation. In: Cole P, Morgan JL, editors. 
Syntax and Semantics, Vol. 3: Speech Acts. New York: Academic Press.

\bibitem{cattuto2} 
Cattuto C, Barrat A, Baldassarri A, Schehr G (2009) 
Collective dynamics of social annotation.
Proc Natl Acad Sci USA {\bf 106}: 10511--10515.

\bibitem{kruskal} 
Kruskal W (1987)
Relative importance by averaging over orders.
Am Stat  {\bf 41}: 6--10.

\end{thebibliography}
\end{document}